\documentclass[]{spie}  %>>> use for US letter paper

\usepackage{amsmath,amsfonts,amssymb}
\usepackage{graphicx}
\usepackage{subcaption}
\usepackage{setspace}
\usepackage{tocloft}
\usepackage{booktabs}
\usepackage{placeins}
\usepackage{float}

\title{Student Beats the Teacher: Deep Neural Networks for Lateral Ventricles Segmentation in Brain MR}

\author[a,c,*]{Mohsen Ghafoorian}
\author[a,d,*]{Jonas Teuwen}
\author[a]{Rashindra Manniesing}
\author[b]{Frank-Erik de Leeuw}
\author[a]{Bram van Ginneken}
\author[a]{Nico Karssemeijer}
\author[a]{Bram Platel}

\affil[a]{Radboud University Medical Center, Diagnostic Image Analysis Group, Department of Radiology and Nuclear Medicine, Nijmegen, the Netherlands}
\affil[b]{Donders Institute for Brain, Cognition and Behaviour, Department of Neurology, Radboud University Medical Center, Nijmegen, the Netherlands}
\affil[c]{TomTom, Amsterdam, the Netherlands}
\affil[d]{Optics Research Group, Imaging Physics Department, Delft University of Technology, the Netherlands}

\authorinfo{Send correspondence to Jonas Teuwen: jonas.teuwen@radboudumc.nl.}

\cftpagenumbersoff{figure}
\cftpagenumbersoff{table} 
\begin{document} 
\maketitle

\begin{abstract}
Ventricular volume and its progression are known to be linked to several brain diseases such as dementia and schizophrenia. Therefore accurate measurement of ventricle volume is vital for longitudinal studies on these disorders, making automated ventricle segmentation algorithms desirable. In the past few years, deep neural networks have shown to outperform the classical models in many imaging domains. However, the success of deep networks is dependent on manually
labeled data sets, which are expensive to acquire especially for higher dimensional data in the medical domain. In this work, we show that deep neural networks can be trained on much-cheaper-to-acquire pseudo-labels (e.g., generated by other automated less accurate methods) and still produce more accurate segmentations compared to the quality of the labels. To show this, we use noisy segmentation labels generated by a conventional region growing algorithm to train a deep network for lateral ventricle segmentation. Then on a large
manually annotated test set, we show that the network significantly outperforms the conventional region growing algorithm which was used to produce the training
labels for the network. Our experiments report a Dice Similarity Coefficient (DSC) of $0.874$ for the trained network compared to $0.754$ for the conventional region growing algorithm ($p < 0.001$).
\end{abstract}

% Include a list of up to six keywords after the abstract

{\noindent \footnotesize\textbf{*}Equal contribution}\\
% Include email contact information for corresponding author
%{\noindent \footnotesize\textbf{\dag}Corresponding author,  \linkable{jonas.teuwen@radboudumc.nl} 
%}

\begin{spacing}{1}   % use double spacing for rest of manuscript

\section{Introduction}
\label{sect:intro}  % \label{} allows reference to this section
Lateral ventricles are anatomical parts of the ventricular system in the brain, where the cerebrospinal fluid is produced. Ventricular volume and its progression are associated with several brain diseases. In certain forms of dementia, the increase of lateral ventricular volume has been associated to decline in cognitive function\cite{haxby1992longitudinal}. Some psychiatric
illnesses such as schizophrenia have also been linked to enlargement in ventricular volume\cite{wright2000meta}. Additionally, asymmetrical shapes between the left and the right lateral ventricles together with the size of the ventricles can be indicative of abnormalities in the brain\cite{McKinney2017}.

Even though a rough estimation of the ventricular volume such as the number of slices that the ventricles appear in, might be sufficient for some applications, more accurate quantitative measurements are necessary to longitudinally study subtle differences. It has also been shown that leveraging spatial information using ventricles as landmarks are beneficial for the detection of a number of pathologies in the brain including white matter hyperintensities \cite{ghafoorian2016automated} and lacunes \cite{ghafoorian2017deep}. Though manual annotation of lateral ventricles might be an option on smaller datasets and cross-sectional studies, this would not be feasible otherwise as the task is time-consuming, laborious and subjective. Therefore an accurate, objective and independent segmentation of the left and right ventricles is desirable in clinical practice. 

With the success of deep neural networks \cite{LeCun2015,Schmidhuber2015} in visual
pattern recognition, many studies have been successfully conducted in the medical image analysis domain during the past few years \cite{LITJENS201760, Teuwen2017}, that have resulted in intelligent systems that reach or surpass the level of medical experts on different tasks and domains \cite{gulshan2016development, ghafoorian2017location, bejnordi2017diagnostic}.

Since the recent deep learning approaches follow a data-driven strategy to learn the optimal representations for the specific tasks at hand, these methods often require large sets of annotated data to train on. Several recent studies have shown strong implications of training dataset size on the quality of trained networks. For instance, it has been shown that even with gigantic datasets, the performance of the trained network linearly scales with logarithm of the size of the training data \cite{Chen17}.

Given the reasoning above, the computer vision community has created enormous labeled datasets using crowd sourcing methods, for instance using Amazon mechanical turk. However this solution is not feasible for medical datasets, as the labeling process requires specific expertise that is only possible with medical experts available. Therefore, the high costs of gathering large medical datasets have still hindered feasibility of gigantic datasets that fully leverage the high capacity of the deep neural networks on various medical image analysis domains. 

Another strategy to provide large labeled datasets is to use (not necessarily very accurate) available methods for the task in order to provide pseudo-labels. Using this, one can provide arbitrarily large datasets as far as unlabeled data is available. This however, arises a few interesting questions to be answered: 1) Considering an imposed trade-off between the dataset size and its relative label accuracy, would that make sense to train neural networks with noisy but large datasets rather than smaller ones with more accurate labels, and 2) In case we opt for the latter, is the low accuracy of the provided pseudo-labels necessarily an upper-bound for the accuracy of a trained network?

In this study, we aim to answer the aforementioned rather important questions by reporting a deep neural network that achieves high accuracy in segmenting the left and right ventricles separately, being trained on noisy pseudo-labels. We also show that, though desirable, accurate manual labels are not mandatory to produce good results, given a large set of (unbiased) noisy-labeled images.

\section{Methods}
\subsection{Material}
The data used in this work is obtained from the RUN DMC\cite{vanNorden2011} (Radboud University
Nijmegen Diffusion Tensor and Magnetic Resonance Imaging Cohort), which is a longitudinal study of small vessel disease and its progression. The imaging protocol includes a 3D T1-magnetization prepared rapid gradient-echo (MPRAGE) pulse sequence with voxel size of $1.0 \times 1.0 \times
1.0~\text{mm}^3$ and a fluid attenuation inversion recovery (FLAIR) pulse sequence with voxel size $0.5 \times 0.5 \times 5.0~\text{mm}^3$ with a slice gap of $1~\text{mm}$, scanned using a 1.5T MR scanner (Magnetom Sonata, Siemens Medical Solution,
Erlangen, Germany).

We selected a subset of 397 subjects which was randomly split into sets of 246, 99 and 52 subjects for training, validation and testing purposes respectively.  

\subsection{Preprocessing}
For an accurate segmentation, we need to take into account the possible movement of the patient between the acquisition of the T1 and FLAIR modalities. To align the image coordinates of both modalities, we rigidly registered the T1 images with the images using FSL-FLIRT\cite{Jenkinson2012}. In order to make the processing easier, we exclude non-brain tissue such as the skull, eyes, etc. We computed the brain mask using FSL-BET\cite{Jenkinson2012} on the T1 images. The resulting masks are then transformed using the computed transformation to the FLAIR images. To correct for the spatial intensity variations on the MR images caused by inhomogeneities in the magnetic field, we perform a bias-field correction using FSL-FAST \cite{Jenkinson2012}. As a final preprocessing step we normalize each image by dividing it with the $95^\text{th}$ percentile of all intensities within the same image.

\subsection{Reference annotations}
To generate the pseudo-labels for the training set, we used an in-house developed application where automatically selected seed points are used to perform a watershed-based segmentation algorithm on the T1 image to provide ventricle masks on the whole training, validation, and test sets. The algorithm is available in the commercial version of MeVisLab (MeVis Medical Solutions AG and Fraunhofer MEVIS; Bremen, Germany). The provided masks generated by the watershed region growing based algorithm are inaccurate in some cases or totally failing in some others. We therefore excluded 9 cases from the training set where the algorithm failed completely. In addition to this for evaluation purposes, the test set was independently manually segmented by an experienced reader on the registered T1 images, where the FLAIR images was used in cases of ambiguity.

\subsection{Network architecture and training procedure}

\begin{figure}[t!]
\begin{center}
\begin{tabular}{c}
\includegraphics[width=0.8\textwidth]{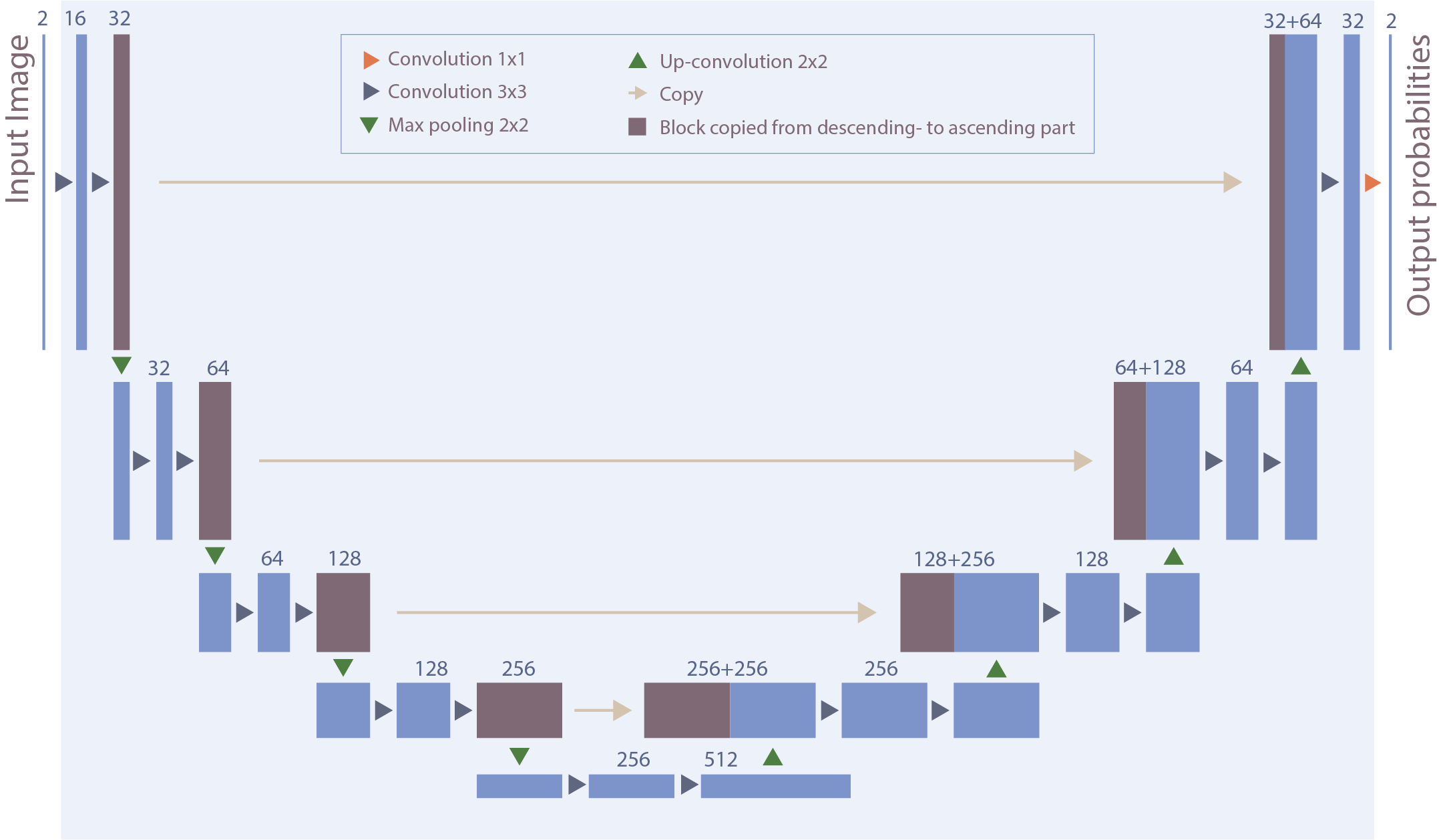}
\end{tabular}
\end{center}
\caption{The fully convolutional network used for training our model. This is a U-net-like architecture \cite{unet,3dunet} with slight modifications as described in the text.} 
\label{fig:unet}
\end{figure} 

To segment the left and the right ventricles separately, we formulated the problem as a three-class segmentation of the background, left ventricle and right ventricle respectively. We utilized a fully convolutional network based on the U-net\cite{unet} architecture, with a depth of $5$, applied slice-by-slice on a two-channel image composed of the T1 and FLAIR modalities. As in the analysis path of the standard U-net, we used $3 \times 3$ convolutional filters and $2 \times 2$ max pooling with $(2, 2)$ stride. We slightly deviated from the original architecture by using leaky ReLus with leakiness $0.01$ and follow these by dropout with $0.1$ probability and a batch normalization \cite{ioffe2015batch} layer. Additionally, we started the first convolutional layer with 16 filters and we already doubled the number of filters in each layer, before the max pooling to avoid bottlenecks\cite{3dunet}. We employed a similar scheme in the synthesis path. Details of the network architecture is illustrated in Figure \ref{fig:unet}.

We used the categorical cross-entropy loss function with $L_2$ regularization with $\lambda_2=10^{-5}$. To account for class imbalance, we weighted the loss function on the background by a factor of $0.01$. The network weights were initialized from a Gaussian distribution
$N(0,{2/\text{fan}_{\text{in}}}$). To train the network, we used the Adam update rule \cite{kingma2014adam} with parameters $\beta_1 = 0.9$, $\beta_2 = 0.999$ and $\epsilon =
10^{-8}$. We trained our network for 200 epochs with an initial learning rate of $10^{-4}$ which was
decreased in later epochs to $10^{-5}$. The final model was selected as the model with the lowest validation loss. The network was trained using the imperfect segmentations made by the region growing method as described in section 2.3.

\begin{figure}[H]
\begin{center}
%\vspace{-0.5cm}
\begin{tabular}{c}
\includegraphics[width=7.5cm]{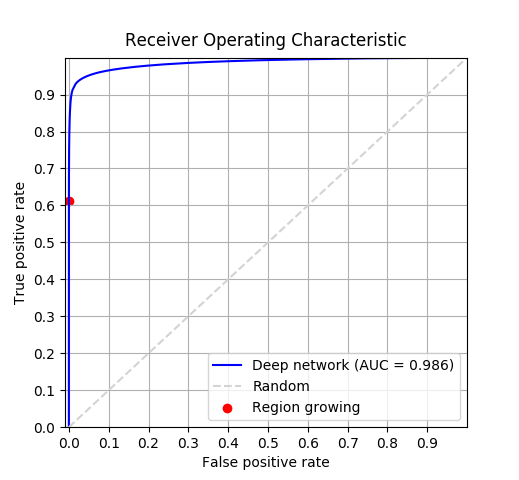}
\end{tabular}
\end{center}
\caption 
{ \label{fig:auc}
Receiver Operating Characteristic for the segmentation of both ventricles. Please note that the region growing algorithm is represented by a single point as the mentioned method is not probabilistic in contrast to the deep network.} 
\end{figure}
\subsection{Evaluation of performance}
We selected a threshold of $0.5$ on the probability maps provided by the deep network to obtain binary segmentations. We compared the classical region growing method to the deep network thresholded output using the manual segmentations as the reference standard with the Dice Similarity Coefficient (DSC). The DSC is given by:
% We selected a threshold of $0.5$ on the probability maps provided by the deep network to obtain binary segmentations. We compared the classical region growing method to the deep network thresholded output using the manual segmentations as the reference standard and the Dice Similarity Coefficient (DSC) as the metric:
\begin{equation*}
\text{DSC}(R, X) = 2\frac{\sum_i |R_i \cap X_i|}{\sum_i |R_i| + |X_i|},
\end{equation*}
where $R_i$ are the reference annotations for subject $i$, $X_i$ are either the labels generated by the conventional approach or the thresholded network outputs and $|\cdot|$ is the size of the set.

We also report and compare receiver operating characteristic (ROC) curves that represent the methods with their true and false positive rates (sensitivity and $1 - $specificity) on various operating points. We use area under the ROC curve (AUC) as a single metric to quantitatively compare ROC curves. Furthermore, we use bootstrapping (over 1000 randomly created bootstraps) on the test set samples to report statistical significance test $p-$values. To be more specific, given ``method A is no better than method B'' as the null-hypothesis to reject, empirical $p-$value is reported as the proportion of bootstraps where method B results in a higher DSC.

\section{Results}
\begin{figure*}[!ht]
\makebox[\linewidth][c]
{
		{\includegraphics[width=.8\textwidth]{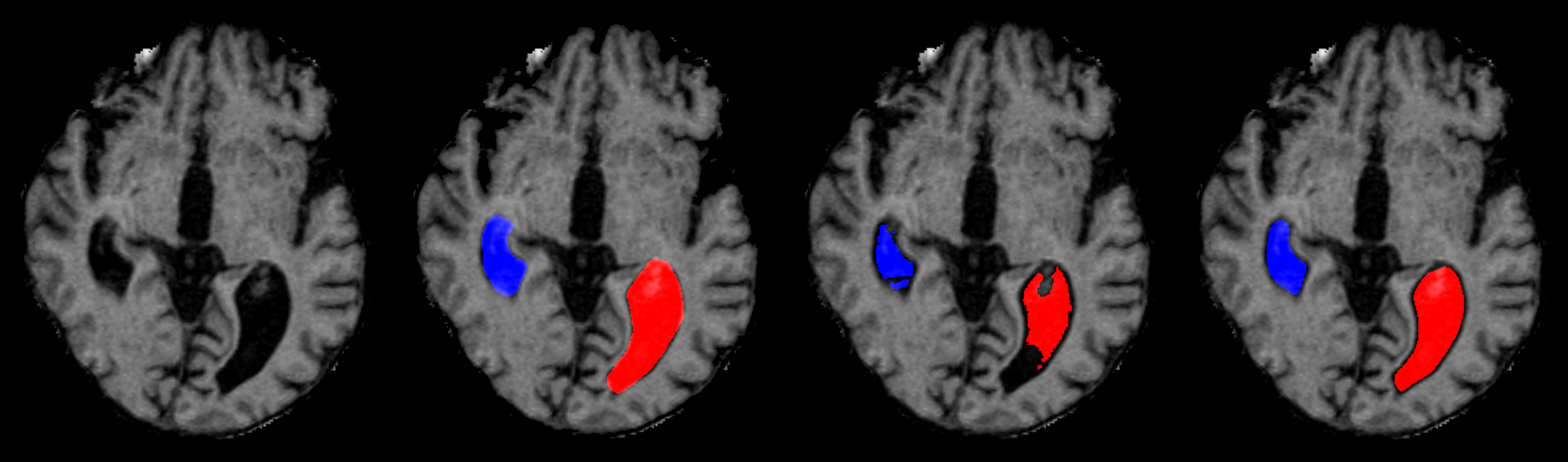}} 
}\\
\makebox[\linewidth][c]
{
		\centering
		{\includegraphics[width=.8\textwidth]{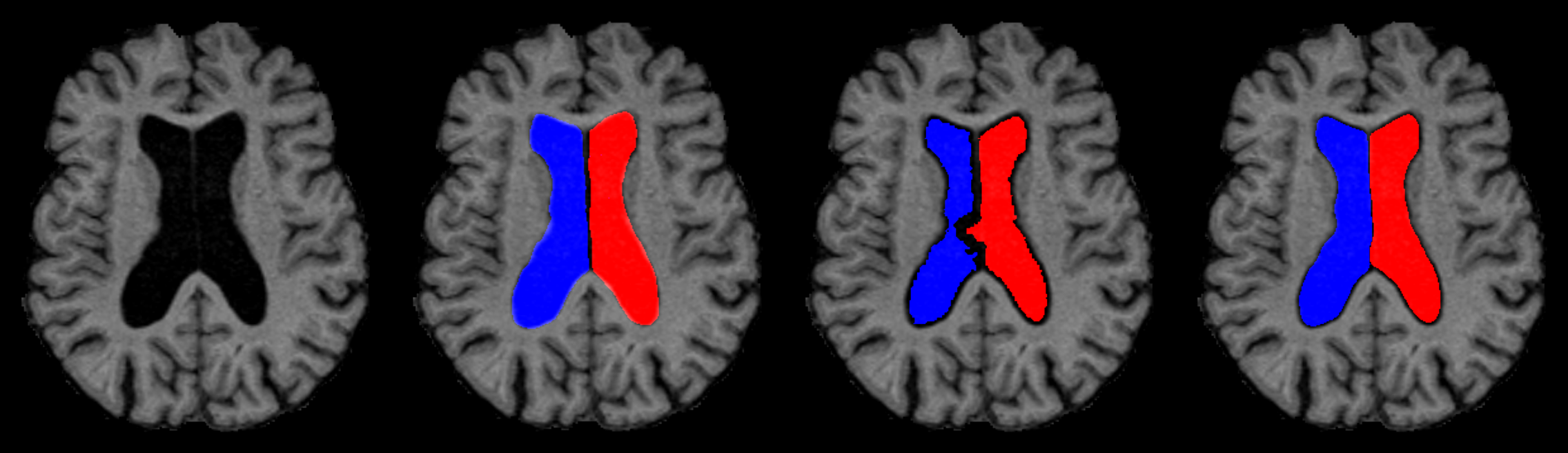}} 
}\\
\makebox[\linewidth][c]
{
		{\includegraphics[width=.8\textwidth]{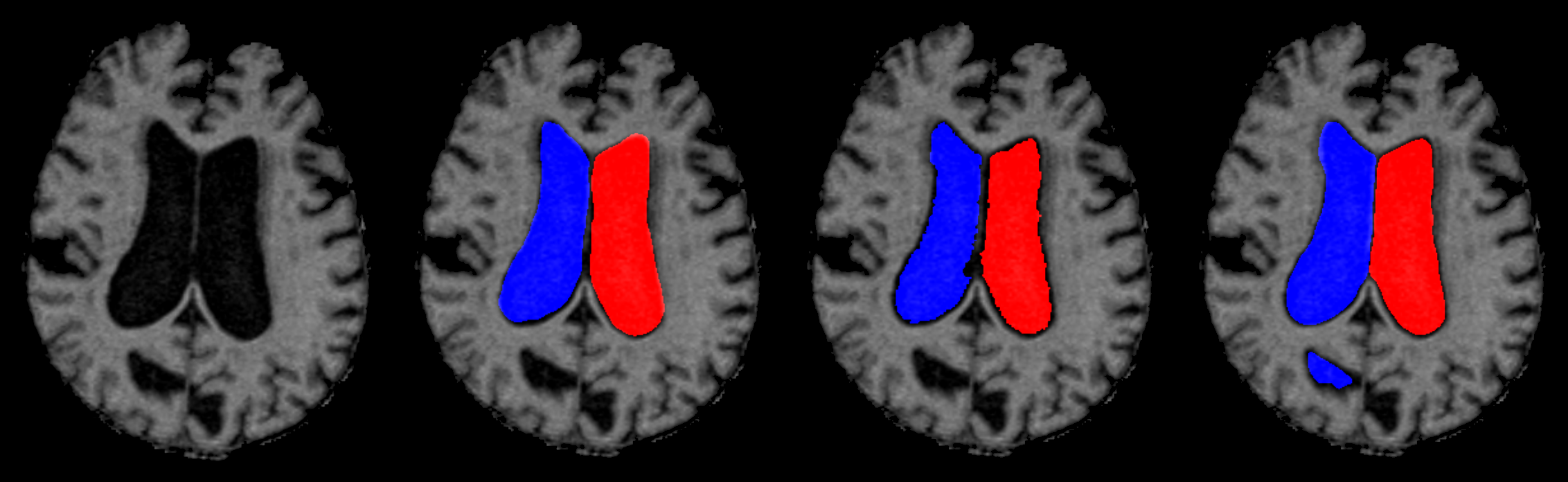}} 
}
\vspace{0.05em}
\caption{Sample slices of qualitatively representing results of our method. The four images in each represent the original T1 image, the manual segmentation, the output of the region growing algorithm used as reference standard and the output of the proposed method respectively. In the first two rows we have given examples where our method clearly outperforms the region growing algorithm. In the last row a case is given where our method makes false positives.}
\label{fig:examples}
\end{figure*}

Evaluating the methods on the test set, we obtained a DSC of $0.874$ compared to $0.754$ for the region growing based method, with respect to the manual annotations as the reference standard. The deep network significantly outperformed the region growing method ($p < 0.001$) on all three comparison scenarios of right, left and both ventricles segmentation, even though the network was trained on the outputs from that method. The DSC for segmentation of left and right ventricles and the whole lateral ventricles for the two methods are presented in Table~\ref{tab:performance}, also shows a consistent improvement over the baseline method. Furthermore, an ROC analysis is illustrated on Figure~\ref{fig:auc} to provide a visual comparison between the two methods. It should be noted that since the region growing based model does not provide probabilistic outputs, its ROC analysis results consists of a single operating point, which is represented by a single point in the ROC graph. In Figure~\ref{fig:examples} we present a qualitative comparison between the different methods on a sample slice.

\begin{table}[h!]
\centering
\caption{Performance of the different methods on the left, right and both
  ventricles respectively.}
\label{tab:performance}
\begin{tabular}{@{}lccc@{}}
\toprule
                          & Left ventricle & Right ventricle & Both ventricles \\ \midrule
  DSC (region growing)      & $0.750$               & $0.723$                & $0.754$                \\
  DSC (deep neural network) & $0.881$              & $0.867$               & $0.874$               \\
  AUC (deep neural network) & $0.990$               &$0.989$                 & $0.986$ \\  \bottomrule
\end{tabular}
\end{table}

\section{Discussion and conclusions}
Interestingly in the experiments, we observed that the deep model trained on imperfect ground truth could still get a decent training and outperform its ground truth generating method significantly. This is an interesting and important finding for the medical imaging domain where the high costs of generating large manually labeled datasets might seem to reject the feasibility of training deep neural networks that require gigantic training sets to achieve good performance. These results also show that a relatively low accuracy of the provided pseudo-labels is not necessarily an upper bound to the performance of the trained network.

For this to happen, there are two requirements that need to be satisfied: Firstly, the distribution of the samples with noisy labels should be adequately randomly scattered over the feature space. Otherwise, if the ground truth providing method is biased and constantly repeats the same error patterns the model would most likely learn the same error patterns. Secondly, the method should be regularized well enough to maintain its generalizability and not to overfit the noise patterns. 

In this work, we presented a fully automated algorithm for the segmentation of the lateral ventricles on brain MR images that is well capable of discriminating between the left and right ventricles. Despite the noisy training labels, the network achieves a DSC of $0.874$.
%The u-net is able to generalize beyond the mistakes made in the ground truth and produces better results than the training data itself. We hypothesize this is due to the fact that the mistakes made in the segmentations of the classical model are reasonable unbiased.  
%\acknowledgments

\bibliography{report}   % bibliography data in report.bib
\bibliographystyle{spiejour}   % makes bibtex use spiejour.bst

\end{spacing}
\end{document}